\documentclass[10pt,twocolumn,letterpaper]{article}

\usepackage[pagenumbers]{cvpr} 
\usepackage{amsmath}
\usepackage{multirow}
\usepackage{booktabs}
\usepackage{amssymb}

\definecolor{cvprblue}{rgb}{0.21,0.49,0.74}
\usepackage[pagebackref,breaklinks,colorlinks,allcolors=cvprblue]{hyperref}

\def\paperID{11739} 
\def\confName{CVPR}
\def\confYear{2025}

\title{DR-BFR: Degradation Representation with Diffusion Models \\ for Blind Face Restoration}

\author{Xinmin Qiu\\
University of Chinese Academy of Sciences\\
Beijing, China\\
{\tt\small qiuxinmin21@mails.ucas.ac.cn}
\and
Bonan Li\\
University of Chinese Academy of Sciences\\
Beijing, China\\
{\tt\small libonan@ucas.ac.cn}
\and
Zicheng Zhang\\
University of Chinese Academy of Sciences\\
Beijing, China\\
{\tt\small zhangzicheng19@mails.ucas.ac.cn}
\and
Congying Han\\
University of Chinese Academy of Sciences\\
Beijing, China\\
{\tt\small hancy@ucas.ac.cn}
\and
Tiande Guo\\
University of Chinese Academy of Sciences\\
Beijing, China\\
{\tt\small tdguo@ucas.ac.cn}
}

\begin{document}
\maketitle
\begin{abstract}
Blind face restoration (BFR) is fundamentally challenged by the extensive range of degradation types and degrees that impact model generalization. Recent advancements in diffusion models have made considerable progress in this field. Nevertheless, a critical limitation is their lack of awareness of specific degradation, leading to potential issues such as unnatural details and inaccurate textures. 
In this paper, we equip diffusion models with the capability to decouple various degradation as a degradation prompt from low-quality (LQ) face images via unsupervised contrastive learning with reconstruction loss, and demonstrate that this capability significantly improves performance, particularly in terms of the naturalness of the restored images. 
Our novel restoration scheme, named \textbf{\textit{DR-BFR}}, guides the denoising of Latent Diffusion Models (LDM) by incorporating Degradation Representation (DR) and content features from LQ images. DR-BFR comprises two modules: 1) \textbf{Degradation Representation Module} (DRM): This module extracts \textit{degradation representation with content-irrelevant} features from LQ faces and estimates a reasonable distribution in the degradation space through contrastive learning and a specially designed LQ reconstruction. 2) \textbf{Latent Diffusion Restoration Module} (LDRM): This module perceives \textit{both degradation features and content features} in the latent space, enabling the restoration of high-quality images from LQ inputs.
Our experiments demonstrate that the proposed DR-BFR significantly outperforms state-of-the-art methods quantitatively and qualitatively across various datasets. The DR effectively distinguishes between various degradations in blind face inverse problems and provides a reasonably powerful prompt to LDM.

\end{abstract}    
\section{Introduction}
\label{sec:intro}

\begin{figure}[t]
\centering
\includegraphics[width=0.9\columnwidth]{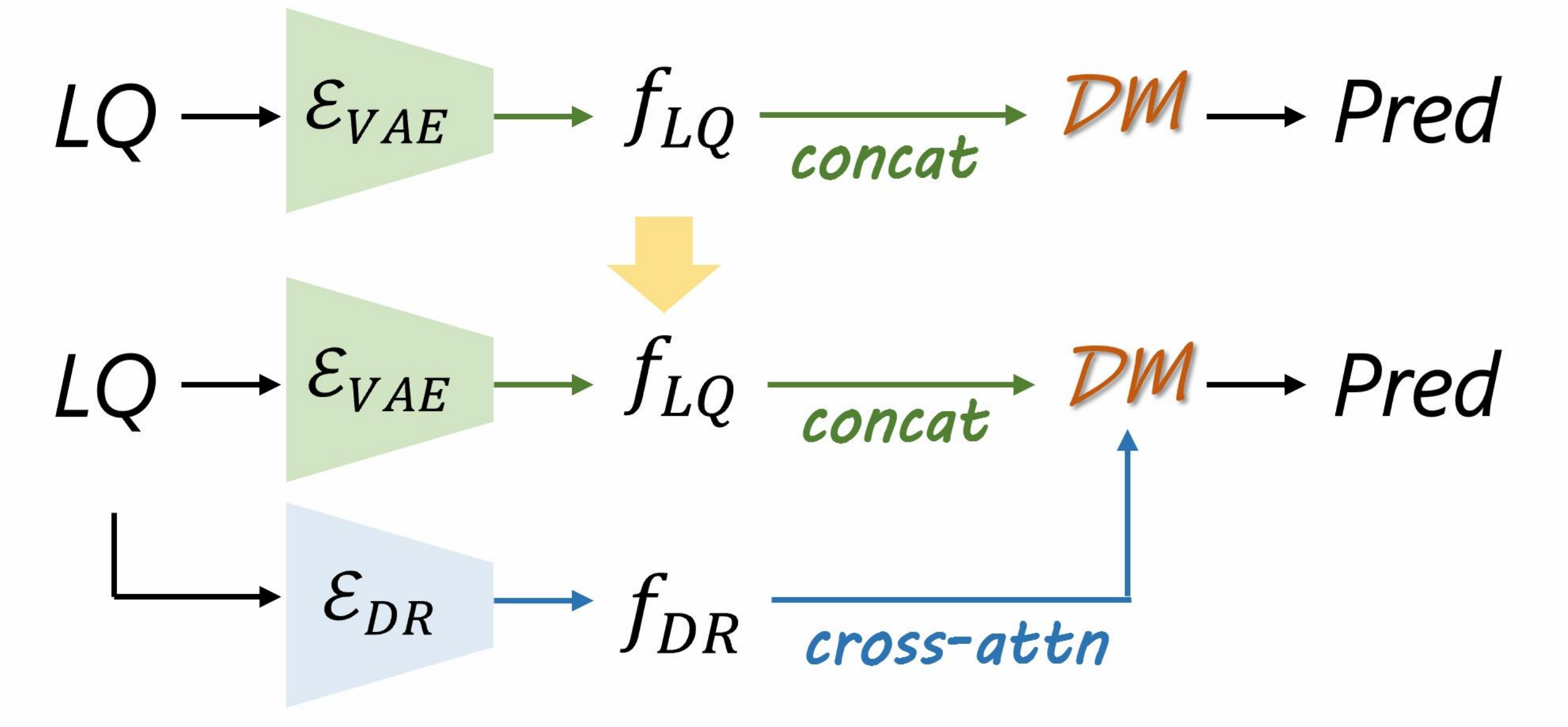} 
\vspace{-3mm}
\caption{Comparison of \textbf{guidance mechanisms } between previous diffusion-based methods and our DR-BFR. The key distinction lies in the approach used to handle LQ face images. In our DR-BFR, the DR features are decoupled in advance.   These decoupled features as a resonably prompt, combined with the extracted content features, serve as guiding conditions for DM.}
\vspace{-6mm}
\label{fig:guidance}
\end{figure}

Blind face restoration (BFR) aims to reconstruct high-quality (HQ) face images from low-quality (LQ) inputs with unknown degradations, holding significant potential for applications such as surveillance image enhancement and the restoration of old photographs. The core challenge in BFR lies in its "blind" nature, where accurately simulating the complex and diverse real-world degradations is inherently difficult. Recent advancements have introduced various generative model solutions. Early approaches were predominantly GAN-based, such as GFPGAN~\cite{GFPGAN} and GPEN~\cite{GPEN}. More recent methods have shifted towards diffusion-based techniques like DiffIR~\cite{diffir}, DR2~\cite{dr2} and others~\cite{DDNM}, which better capture real data distributions and surpass GAN-based methods, particularly in terms of perceptual quality.

While diffusion-based methods have advanced this field, they often excel only with specific degradation types and struggle with real-world images. These models typically fail to clearly distinguish between different types and degrees of degradation, leading to confusion between degradation artifacts and original textures in LQ inputs. 
In particular, diffusion-based methods only employ the LQ image as the conditional guidance previously, and the complexity of the distribution of the guidance has a great influence on the restoration effect of the diffusion model (DM). Blind problem due to various degenerate random combinations, the extracted condition distribution is complicated, and it is impossible to point out useful information points.
Previous approaches, such as DASR~\cite{dasr, dasr2} and other degradation-aware (DA) methods, have enhanced the generalization ability of super-resolution models by pre-estimating degradation features from the input image. Building on this concept, we recognize that \textit{pre-decoupling the degraded representation and incorporating it into Diffusion Models} is crucial for enhancing the generalization and restoration accuracy of these models, while the topic remains unexplored. Therefore, in this paper, we take the first step in extracting content-independent degraded representations (DR) that encapsulate essential degradation information from LQ images to improve diffusion-based methods.

This paper introduces DR-BFR, an innovative approach that capitalizes on DR as a \textbf{\textit{degradation prompt}} to improve the accuracy of restoration in the BFR task within DM. Previous studies~\cite{allinone, imagen, promptrr, prompt_seek, prompt_sr, prompt_ucip, promptrestorer} have substantiated that fine-grained prompts result in fine-grained outcomes, as they offer vital contextual information that enhances the concentration of the target distribution and mitigates uncertainty during both training and sampling phases. In BFR, while LQ images could be considered as \textbf{\textit{visual prompts}} to guide the content restoration process, diffusion-based methods yet struggle to capture intricate details due to the heterogeneous nature of degradation types. 
Previous methodologies have frequently utilized advanced encoders to achieve precise extraction of content features from intricate low-quality samples. In contrast, our approach employs a more streamlined method, emphasizing the extraction of a more coherent representation via regression techniques. This strategy involves directing the DM to execute subtraction operations and harnessing the substantial impact of prompts within the diffusion-based framework.
Consequently, by utilizing DR as a degradation prompt, DM can more effectively exploit degradation information and effectively learn reasonable generation pathways. DR is an implicit feature representation, and offers greater expressive power compared to traditional prompts, particularly in the context of BFR.
This integration of DR and DM significantly increases the restoration precision of LQ images, especially when dealing with limited training data.

Technically, we first extend this method of training DR encoders by combining contrast learning and reconfiguration, modifying the network to be more suitable for multiple degenerate random combination tasks, like BFR. Internally, two patches are randomly chosen from each LQ image, and these patches have nearly identical degradation but distinct content. The DR extracted from patches of the same LQ image serve as positive samples for each other. Externally, given the extensive parameter space formed by various degradation types and degrees, the degradation parameters for each image pair in the same batch are different. The DR extracted from different LQ images act as negative samples for each other.  
Furthermore, we incorporate a supervised training objective to enhance the learned DR features, where a generative model takes the extracted DR and the HQ image to reconstruct the LQ image. This makes the DR independent of image content while maximizing degradation information.
Finally, as illustrated in Figure~\ref{fig:guidance},  we utilize DR to guide the DM by employing a cross-attention mechanism, with an additional adapt coefficient linked to the number of steps $t$.
This adaptation enables the DM to effectively leverage the degradation information supplied by the DR, thereby enhancing restoration accuracy and efficiency. 
This insertion aims to reduce the complexity for the diffusion model when handling LQ images with diverse degradations by pre-decoupling degradation from content. Furthermore, introducing varying degrees of degradation cues at different stages of the inverse diffusion process facilitates a more natural and coherent generation process.


\begin{figure*}[t]
    \centering
\includegraphics[width=1.0\linewidth]{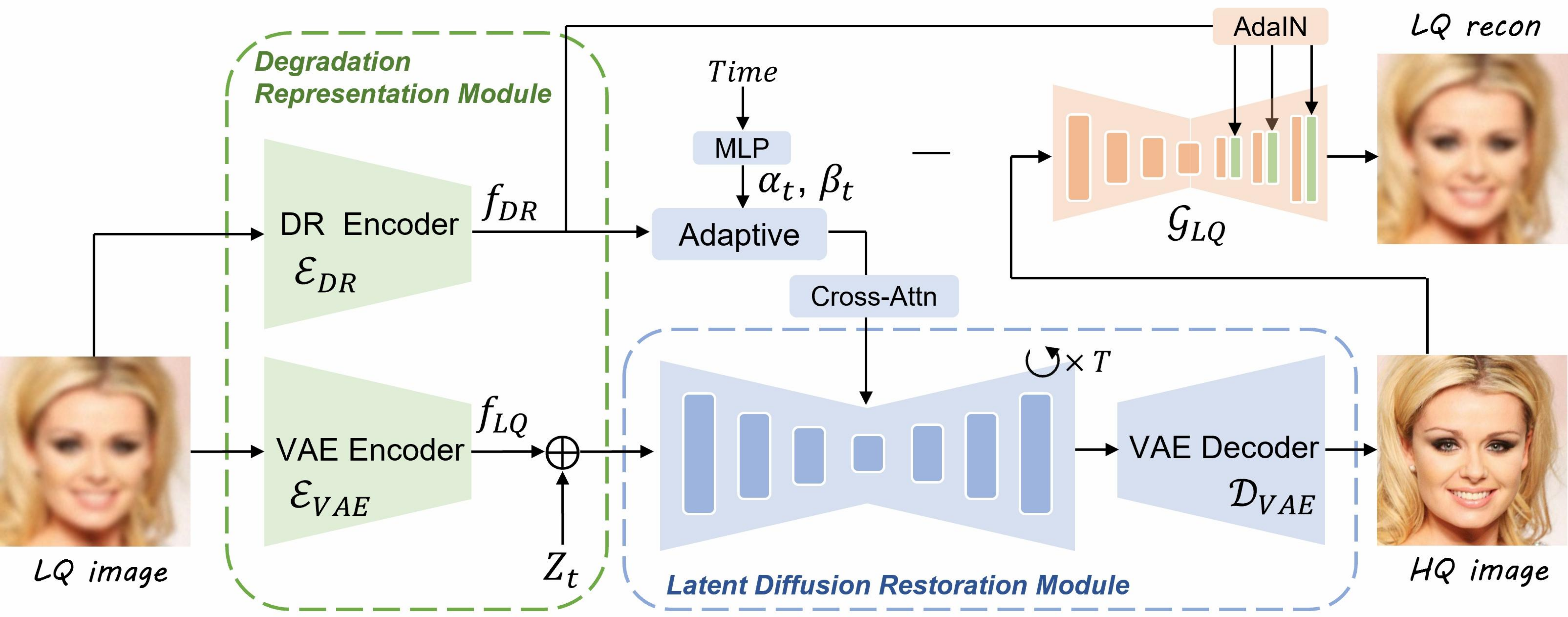} 
\vspace{-6mm}
\caption{\textbf{Framework of the proposed \textit{DR-BFR}} for blind face restoration task. DR-BFR is fundamentally a multi-input conditional diffusion model with specific guidance. Given the LQ face, \textit{DRM} decouples the content-irrelevant degradation information, while an encoder, inherited from LDRM, extracts the content of the LQ image. The diffusion-based design of DR-BFR, which incorporates degradation awareness, demonstrates superior performance as validated by experimental evidence.}
\vspace{-6mm}
\label{fig:framework}
\end{figure*}

\noindent
Our main \textbf{contributions} are summarized as follows:
\begin{itemize}
\item 1) We propose DR-BFR, leveraging \textit{DR features and content features to guide the DM in the same latent space} to restore LQ face with unknown degradation, where the core is to utilize DR as the learnable degradation prompt. Compared to enhancing the complexity of DM, our insight is to strengthen the expressive power of the conditional prompt, providing a simple but effective solution.
\item 2) The training of DRM integrates contrastive learning and LQ reconstruction, utilizing known distribution expressions to accurately adjust the distance between DR features. This ensures that the resulting DR feature is content-independent and comprehensively represents all degradation information.
\item 3) We concatenate LQ content features, use cross-attention for DR features and combine the number of steps $t$ to derive the current adapt coefficient. This conditional guidance enhances DM more natural and faithful to restore LQ face images with unknown degradation.
\item 4) On both synthetic and real-world LQ face datasets, DR-BFR outperforms SOTA methods, particularly in naturalness and fidelity metrics.
\end{itemize}

\section{Related Work}
\noindent\textbf{\textit{{Blind Face Restoration.}}}
Blind face restoration~\cite{DFDNet, GFPGAN, Restoreformer, pgdiff} is a crucial subfield within image restoration. In recent years, significant advancements have been made in BFR task, encompassing various methodological approaches such as CNN-based~\cite{PULSE, DFDNet}, GAN-based~\cite{GPEN, GFPGAN, PSFRGAN}, dictionary-based~\cite{VQFR, Codeformer, Restoreformer}, and diffusion-based~\cite{DifFace, dr2} methods. GFPGAN \cite{GFPGAN} and GPEN \cite{GPEN} leverage GAN-based generative models within an encoder-decoder framework to embed face prior. RestoreFormer~\cite{Restoreformer} integrates classical dictionary-based methods with contemporary vector quantization (VQ) techniques~\cite{vqgan}. Besides, DiffBIR~\cite{diffbir} enhances prior knowledge by employing pre-trained stable diffusion as generation priors. PGDiff~\cite{pgdiff} models desired properties of HQ images, such as structure and color statistical properties.

\noindent\textbf{\textit{{Diffusion Models in Image Restoration.}}}
The diffusion model~\cite{DDPM, DDIM, IprovedDDPM, ldm, InDI, selective, res_shift} has demonstrated exceptional performance in the generative field, producing HQ samples with more accurate distribution and greater training stability compared to GANs~\cite{gannet, styleswinGAN, GAN_I2I}. Owing to its generative priors, diffusion models have been employed in various low-level image tasks~\cite{SR3, diffir}. DDNM~\cite{DDNM} proposes a unified theoretical framework for arbitrary linear image restoration (IR) problems. DDRM~\cite{DDRM} introduces a variational inference objective to learn the posterior distribution of the inverse problem at hand. In this work, we utilize Latent Diffusion Models (LDM)~\cite{ldm} as the foundational framework and design VQVAE~\cite{vqgan} that maps to the latent space to calculate diffusion and reverse diffusion, which effectively reduces the computation complexity.

\noindent\textbf{\textit{{Degradation Representation Learning.}}}
Recent studies~\cite{AirNet, NDR, dasr2, o2v_da, deflow_da, Implicit_da, decodable_da} have increasingly utilized neural networks to directly learn degenerate representations. DASR~\cite{dasr} introduces an unsupervised learning scheme for blind super-resolution (SR) without explicit degradation estimation. ReDSR~\cite{ReDSR} extracts degradation information from low-resolution (LR) images by learning DR to guide the degrader in reproducing the input LR images. DaAIR~\cite{DaAIR} proposes a dynamic path that uniquely associates degradations to explicit experts. 
Inspired by ReDSR~\cite{ReDSR}, we propose a novel approach that integrates unsupervised contrastive learning with supervised LQ reconstruction to solve complex and unknown cases of degenerate combinations. By processing batches of images with various degradations, our method aims to obtain feature representations that are image content-irrelevant while maximizing the extraction of degradation information.
\section{Methodology}
\vspace{-2mm}
The \textit{core idea} of our approach lies in obtaining a degraded representation (DR) feature from LQ images that are independent of the image content, effectively serving as a degradation prompt reasonably. Along with LQ images, this DR feature provides \textbf{\textit{two distinct conditions}} to guide the restoration model in generating more accurate and natural HQ face images. As illustrated in Figure~\ref{fig:framework}, $f_{DR}$ denotes the degradation prompt that is content-irrelevant, and $f_{LQ}$ represents the visual prompt.

\vspace{-2mm}
\subsection{Degradation Representation Module}
\begin{figure}[t]
    \centering
    \includegraphics[width=1.0\linewidth]{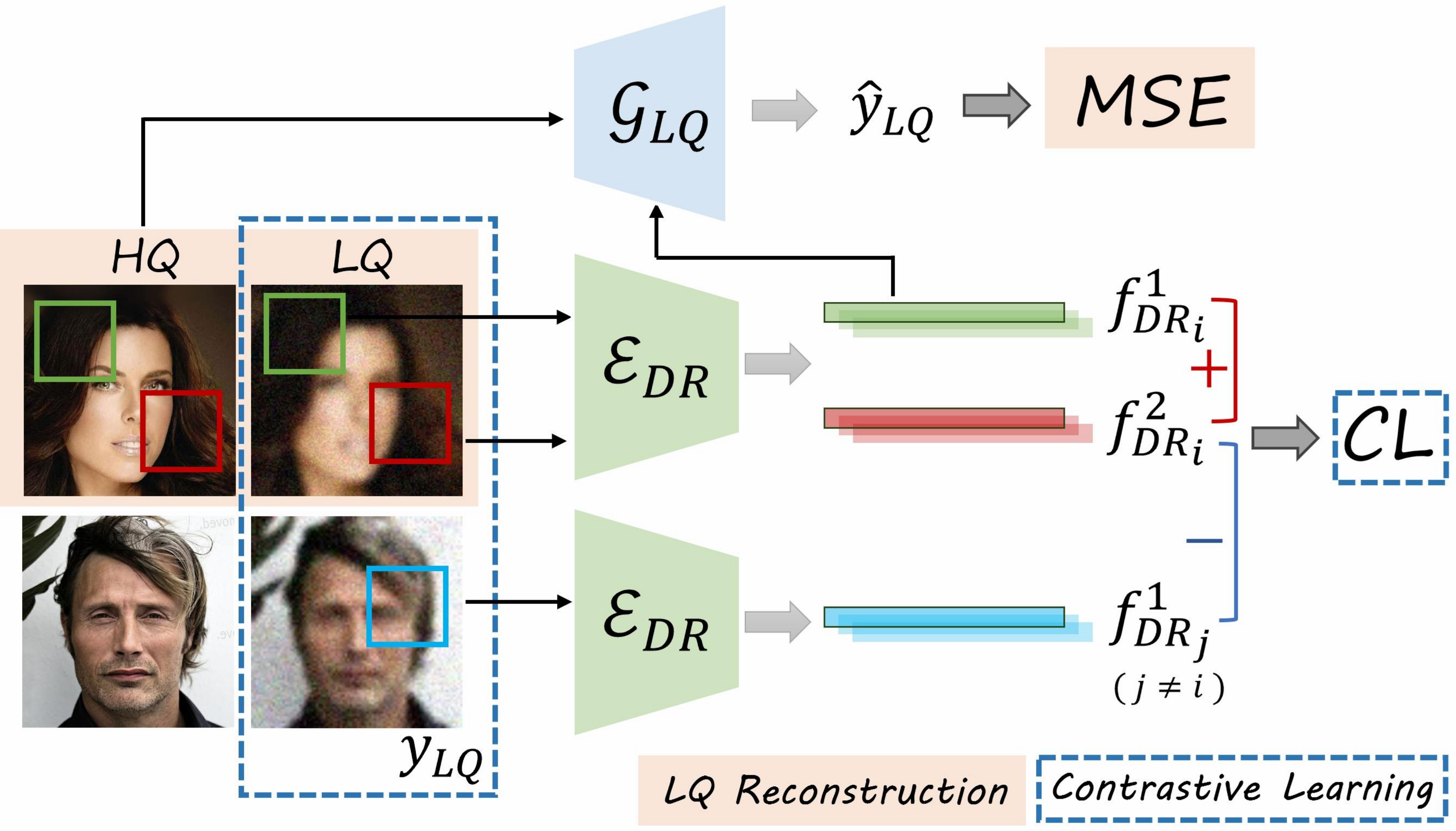}
    \vspace{-4mm}
    \caption{\textbf{Training graph of the Degradation Representation Module}. Here, $\hat{y}_{LQ}$ represents the LQ image generated by the reconstruction of DR and HQ image. Red brackets indicate positive samples within the same batch, while blue brackets denote negative samples.}
    \vspace{-6mm}
    \label{fig:DR_frame}
\end{figure}

The degraded representation module (DRM) trains the feature extractor $\mathcal{E}_{DR}$ using contrastive learning. This module is capable of reconstructing the degraded features and HQ images into LQ images, ensuring the rational distribution of the degraded representation. As depicted in Figure~\ref{fig:DR_frame}, we employ a patch contrastive learning approach, positing that \textit{the DR of different patches within the same image are nearly consistent, despite variations in image content}. Additionally, we add a LQ image generation module, denoted as $\mathcal{G}_{LQ}$, to facilitate the training of the feature extractor. This module takes DR features and HQ images as inputs and reconstructs the corresponding LQ images as outputs. 
To generate degenerate styles of random composition such as blur, compression, etc., we utilize UNet structures.
Moreover, inspired by ReDSR~\cite{ReDSR}, we align the obtained degraded feature distribution with a known distribution, thereby ensuring a reasonably arrangement of the degraded distribution.

Specifically, we denote a batch of HQ images as $\{x_i\}_{i=1}^m$, and LQ images as $\{y_i\}_{i=1}^m$  generated by:
\begin{equation}
y_i=Degradation(x_i;\phi_i)\in \mathbb{R}^{H\times W \times 3},
\label{formula_degradation_ori}
\end{equation}
where $\phi_i\ (i\in \{1,...,m\})$ refers to a random degradation factor, as in Eq.\eqref{formula_degradation}. For each LQ-HQ pair, patches at the same position are randomly extracted separately:
\begin{equation}
    \begin{aligned}
        (r_i^1, r_i^2), (p_i^1, p_i^2)\ &= RandomPatch(x_i, y_i),
    \end{aligned}
    \label{formula_patch}
\end{equation}
where $(r_i^1, r_i^2)$ denote the HQ patches, while $(p_i^1, p_i^2)$ denote the LQ patches located at the same position.

An encoder $\mathcal{E}_{DR}$ is set up to obtain the degenerate representation. It consists of a ConvNet-based network with multiple layers of 3$\times$3 kernel and pooling layers, analogous to the VGG-19~\cite{vgg}. We formulate the process as  
\begin{equation}
    \begin{aligned}
        f^1_{DR_i}=\mathcal{E}_{DR}(p_i^1),\quad f^2_{DR_i}=\mathcal{E}_{DR}(p_i^2)\in \mathbb{R}^{d\times l},
    \end{aligned}
    \label{formula_patch}
\end{equation}
where $d$ and $l$ denote the feature dimensions, respectively.

For the LQ-HQ pair $(x_i,y_i)$,
we consider that different patches on the same LQ image have almost the same degradation, while their contents are different. 
Therefore, we combine different degenerate representations and HQ patches to reconstruct LQ patches. The generator $\mathcal{G}_{LQ}$ utilizes a U-Net structure with Adaptive Instance Normalization (AdaIN) to inject the DR features into the decoder:
\begin{equation}
    \begin{aligned}
\hat{p}_i^2&=\mathcal{G}_{LQ}(r_i^2, f^1_{DR_i}),\\
\mathcal{L}_{recon}&=\sum_i \parallel \hat{p}_i^2 - p_i^2 \parallel.
    \end{aligned}
    \label{formula_recons_LQ}
\end{equation}
Moreover, to make full use of the degradation information in the batch, we construct unsupervised contrastive learning with temperature parameter $\tau$ on the LQ patches:
\begin{equation}
    \mathcal{L}_{contras} = \sum_i -\log \frac{q_i k_i^+}{\sum_{j\not= i}q_i k^-_j/\tau},
\end{equation}
\begin{equation}\notag
\small
    \begin{aligned}
       q_i = MLP(f^2_{DR_i}), k_i^+ = MLP(f^1_{DR_i}), k_j^- = MLP(f^1_{DR_j}),
    \end{aligned}
    \label{formula_contrastive} 
\end{equation}
To ensure a reasonable arrangement of the degraded distribution, we take samples $\{t_i\in \mathbb{R}^{d\times l}\}_{i=1}^m$ from a Gaussian distribution to construct the distribution loss:
\begin{equation}
    \begin{aligned}
        \mathcal{L}_{distribution} &= \frac{2}{m^2}\sum_{i,j=1}^m \parallel f_{DR_i}^1 - t_j\parallel \\
        - \frac{1}{m^2}\sum_{i,j=1}^m \parallel &f_{DR_i}^1 - f_{DR_j}^1\parallel  - \frac{1}{m^2}\sum_{i,j=1}^m \parallel t_i - t_j\parallel.
    \end{aligned}
\end{equation}

Finally, the training loss function for the degenerate representation is as follows:
\begin{equation}
    \begin{aligned}
        \mathcal{L}_{DR}=\mathcal{L}_{recon} + \lambda_1\mathcal{L}_{contras} + \lambda_2 \mathcal{L}_{distribution},
    \end{aligned}
    \label{formula_loss_DR}
\end{equation}
where $\lambda_1$ and $\lambda_2$ represent the weight coefficients of contrastive learning and distribution construct. Notably, we use $\mathcal{E}_{DR}$ to extract degenerate representation features during inference, while $\mathcal{G}_{LQ}$ only plays an auxiliary role in training.

\begin{figure*}[t]
    \centering
    \includegraphics[width=1.0\linewidth]{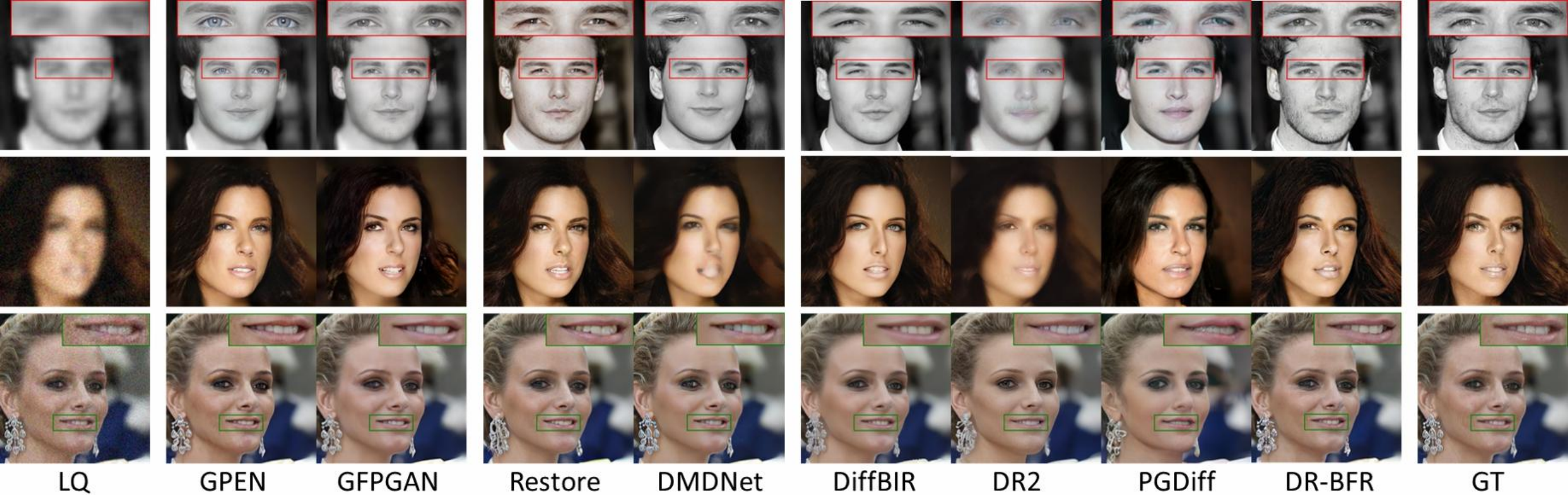}
    \vspace{-6mm}
    \caption{\textbf{Qualitative comparisons on the CelebA-Test} for blind face restoration. Our DR-BFR demonstrates strong performance in detail enhancement, hue preservation and attitude preservation compared to these latest GAN-based, dictionary-based, and diffusion-based methods. Restore denotes RestoreFormer. \textbf{Zoom in for best view}.}
    \vspace{-4mm}
    \label{fig:celebA}
\end{figure*}

\begin{table*}[t]
\setlength{\abovecaptionskip}{0.cm}
\setlength{\belowcaptionskip}{-0.cm}
\centering
\setlength\tabcolsep{4pt}
\begin{tabular}{c|c|cccccccccc|c}
\bottomrule
\multirow{3}*{Metrics} & \multirow{3}*{Input} & \multicolumn{10}{|c}{Methods}\\
\cline{3-13}
~ & ~                     & GPEN  & GFP   & VQFR  & CodeFor &Restore& DMDNet &DifFace&DiffBIR& DR2 & PGDiff &\multirow{2}*{\textbf{DR-BFR}}\\
~ & ~ & \textit{CVPR} & \textit{CVPR} & \textit{ECCV} & \textit{NIPS} & \textit{CVPR}& \textit{TPAMI} & \textit{Arxiv} & \textit{Arxiv} & \textit{CVPR} & \textit{NIPS} &\\
\hline
SSIM$\uparrow$    & 0.6460&	\underline{0.6777}&	\textbf{0.6827}&	0.6382&	0.6494&	0.6219&	0.6727&0.6494&	0.6570&	0.6565& 0.6220 & 0.6429\\
PSNR$\uparrow$    & 24.921&	\textbf{25.423}&	\underline{25.401}&	23.568&	24.907&	24.206& 25.3182&	24.055&	25.297&	24.214& 22.921 & 24.549\\
\hline
FID$\downarrow$   & 93.564& 22.508& 20.676&	17.863&	\underline{16.305}& 17.080&	22.789& 19.654& 19.288&	32.627& 22.547 & \textbf{13.686}\\
NIQE$\downarrow$  & 9.1407& 6.7775&	6.7324&	5.9606&	5.9168&	\underline{5.3440}&	6.7038& 6.1638&	6.4053&	8.1750& 5.4556 & \textbf{5.0113}\\
LPIPS$\downarrow$ & 0.5953&	0.2956&	0.2823&	0.2616& \textbf{0.2441}&	0.2702&	0.2965& 0.3052&	0.2689& 0.3460& 0.3011 & \underline{0.2499}\\
\toprule
\end{tabular}
\caption{\textbf{Quatitative comparison on CelebA-Test} with 3000 randomly selected images for blind face restoration. \textbf{Bold} and \underline{underline} indicate the best and the second best performance. Our DR-BFR excels in naturalness and perceptual metrics. GFP denotes GFPGAN, CodeFor denotes Codeformer and  Restore denotes RestoreFormer.}
\vspace{-4mm}
\label{tab:celeba_results}
\end{table*}

\subsection{Latent Diffusion Restoration Module}
We employ the Latent Diffusion Model (LDM)~\cite{ldm} as the foundation framework for image restoration, referred to as LDRM. To integrate LQ image content and DR features in the diffusion process, we utilize concatenation and a cross-attention mechanism at the input of the estimation model UNet to learn the conditional guidance from these two modalities. 

On the one hand, LDM, as a generative model, can generate clean data that conforms to the distribution of HQ face images from pure Gaussian noise. This generation process is unconditional, and resulted clean images are obtained arbitrarily, aligning with the objective of achieving clear and natural textures. On the other hand, we propose a \textbf{\textit{suitable multi-modal guidance condition}} to achieve faithful restoration results, which ensures that the restored image remains consistent with the original LQ image content, including identity and all the details pertinent to person recognition.

The role of LQ feature $f_{LQ}$ as visual prompt is to guide the generation process in preserving content information. The role of DR features $f_{DR}$ as degradation prompt is to facilitate the decoupling of degradation information, thereby reducing the complexity of generative models faced by blind inverse problems. The training loss for LDM is as follows: 
\begin{equation}
    \begin{aligned}
    \mathcal{L}_{LDM} &= \parallel \epsilon-\epsilon_\theta(z_t,f_{LQ},f_{DR},t)\parallel,\\
    f_{LQ} &= \mathcal{E}_{VAE}(y),\quad
    f_{DR} = \mathcal{E}_{DR}(y).
    \label{formula_ldm}
    \end{aligned}
\end{equation}

In the context of generative model, the generation under the conditional guidance represents the manifestation of conditional probability, enabling effective prediction of inverse problems when the degradation is known. However, for blind inverse problems with unknown degradation, relying solely on single conditional guidance results in a less concentrated distribution of conditional probability. The generative model not only needs to extract content features unrelated to degradation from the LQ image, but also distinguish between content features under varying degradation conditions.  

This decoupling process poses significant challenges for the generative model. To address this, we preprocess the conditional probability of the inverse diffusion process as follows:
\vspace{-2mm}
\begin{equation}
    \begin{aligned}
        P_\theta(z_{t-1}|z_t, f_{LQ}) \Rightarrow P_\theta(z_{t-1}|z_t, f_{LQ}, f_{DR}),
    \end{aligned}
\end{equation}
where $\theta$ represents the model parameters of UNet.

It is difficult to extract content features that are \textit{independent of degradation, while it is relatively easier to extract content features that are independent of degradation}. We compare two methods of image conditional insertion, including cross-attention and concatenation, to determine which insertion method can better guide the image generation process to obtain more accurate and natural results.   
\begin{equation}
    \begin{aligned}
    CrossAttention(\alpha_t\cdot f_{DR}+\beta_t, z_t),\\
    \alpha_t,\beta_t = MLP(t).
    \label{formula_DR_inject}
    \end{aligned}
\end{equation}

\begin{figure*}[t]
    \centering
    \includegraphics[width=1.0\linewidth]{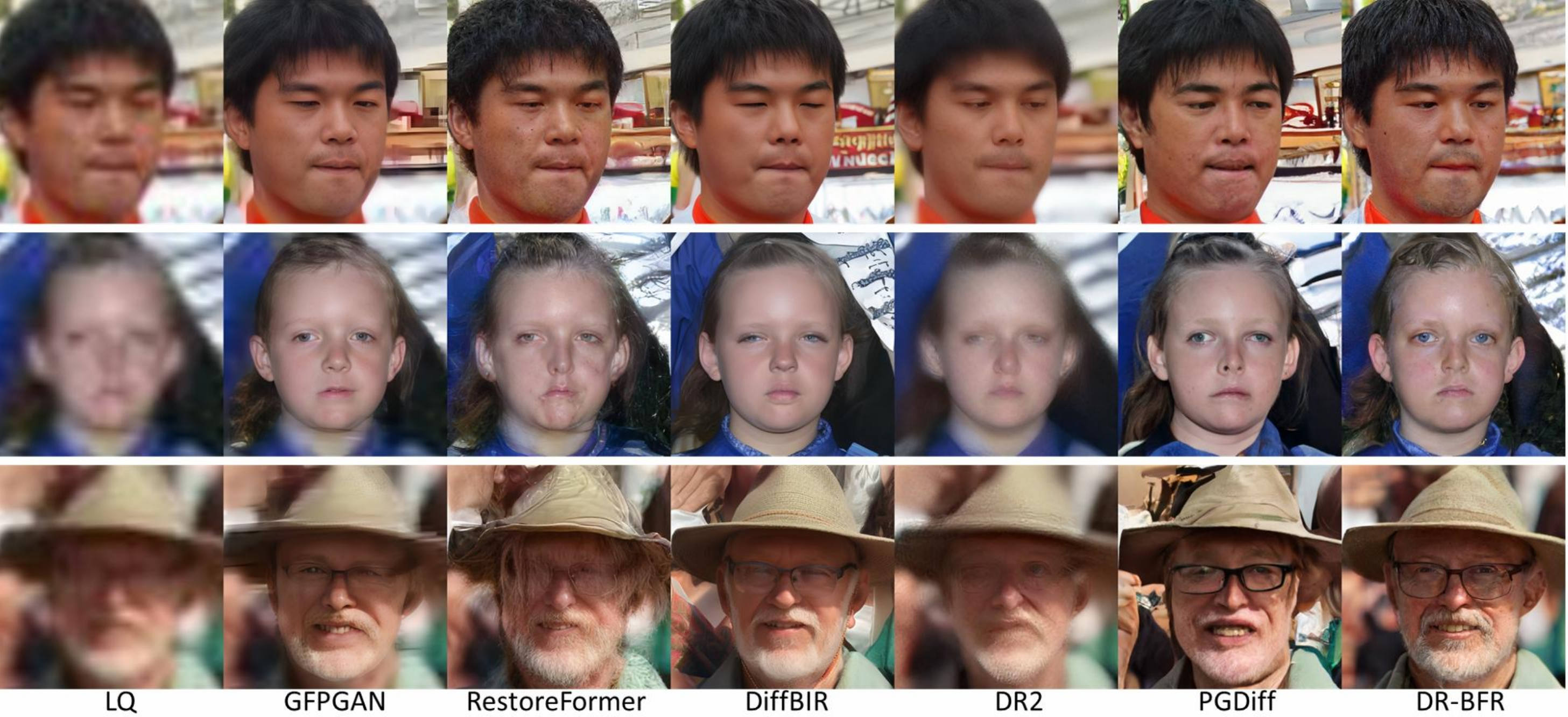}
    \vspace{-6mm}
    \caption{\textbf{Qualitative comparisons on real-world datasets}. Our DR-BFR demonstrates superior performance in both detail enhancement and hue preservation, particularly on inputs with severe degradation. \textbf{Zoom in for best view}.}
    \vspace{-3mm}
    \label{fig:real}
\end{figure*}
\begin{table*}[t]
\setlength{\abovecaptionskip}{0.cm}
\setlength{\belowcaptionskip}{-0.cm}
    \begin{center}
    \setlength\tabcolsep{6.8pt}
    \begin{tabular}{lc|cc|cc|cc|cc}
    \bottomrule
    Dataset & \multirow{2}*{Type-based}& \multicolumn{2}{|c}{LFW} & \multicolumn{2}{|c}{CelebChild} & \multicolumn{2}{|c}{WebPhoto-Test} & \multicolumn{2}{|c}{Wider}\\
    Method &~ & FID$\downarrow$ & NIQE$\downarrow$ & FID$\downarrow$ & NIQE$\downarrow$ & FID$\downarrow$ & NIQE$\downarrow$& FID$\downarrow$ & NIQE$\downarrow$ \\
    \hline
     LQ                     & - & 6.4582&	94.03&	6.6542&	119.48&10.0806&155.91&12.0099&222.95\\
     GPEN               &GAN & 6.0327&	46.15&	6.6999&	\underline{105.85}&	6.6234& 68.22& 6.2069& 32.54\\
     GFPGAN             &GAN & 6.2330&	42.97&	7.0313&	114.47&	7.1427& 80.39& 6.9093& 43.72\\
     VQFR          &Codebook & 5.5092&	41.80&	6.5187&	120.21&	6.1202& 63.38& 5.4927& \textbf{28.18}\\
     CodeFormer    &Codebook & 5.3945&	42.68&	6.4258&	114.06&	5.8507&	64.76& 5.2382& \underline{29.40}\\
     RestoreFormer &Codebook & \underline{4.9487}&	41.38&	\underline{5.5914}&	109.52&	\underline{5.2526}&	61.65& \underline{4.6154}& 38.29\\
     DMDNet        &Codebook & 5.5569&	43.28&	6.4132&	109.59&	6.7571&	66.65& 6.0360& 33.76\\
     DifFace      &Diffusion & 6.8339&	54.64&	6.7321&	113.56&	7.3431&	78.39&	7.9364&	59.70\\
     DiffBIR      &Diffusion & 5.8180&	\textbf{39.11}&	6.2220& 114.18& 6.3345&	68.08&	5.6994&	25.27\\
     DR2          &Diffusion & 6.2578&	\textit{40.99}&	7.0488&	123.63&	8.2475&	95.60&	8.6105&	59.55\\
     PGDiff       &Diffusion &   5.1049    &  \underline{40.14}   &	5.8718&	109.01&	5.3651&	\underline{61.35}&  4.9502&	31.20\\
     DR-BFR      &Diffusion  & \textbf{4.8198}&	41.67&	\textbf{5.3116}&	\textbf{104.95}&	\textbf{5.0806}&	\textbf{61.20}&	\textbf{4.5283}&	30.08\\
    \toprule
    \end{tabular}
    \end{center}
    \vspace{-2mm}
    \caption{\textbf{Quatitative comparison on real-world datasets}. \textbf{Bold} and \underline{underline} indicate the best and the second best performance. The non-reference metric NIQE assesses the quality of restoration, while the distribution gap between FFHQ and results is quantified by FID.}
    \vspace{-6mm}
    \label{tab:realworld}
\end{table*}

In contrast, the insertion of degradation representation features is inserted in a similar way to prompt, and the step $t$ is used as the adapt coefficient, as shown in (\ref{formula_DR_inject}). The purpose is to provide the network with different degradation representation degrees at different steps, so as to promote the guided restoration process to be more natural and controllable.
\section{Experiments}

\noindent{\textit{\textbf{Training Datasets.}}}
We choose FFHQ~\cite{styleGAN} as the training dataset, which contains 70,000 high-quality PNG format face images.
Since our DR-BFR relies on supervised training, the corresponding LQ-HQ image pairs are required. To simulate LQ images in real-world, we employ a degradation model generated through random sampling. Its generation formula~\cite{degradation2, GFPGAN} is presented in Eq.\eqref{formula_degradation}, where $y$ is the HQ image, $k_{\sigma}$ is the Gaussian blur kernel, $r$ represents the down-sampling scale factor, and $q$ represents the JPEG compression of the image with quality factor $q$. To ensure direct comparability with the experimental results of recent BFR methods, we randomly sample the parameters $\sigma$, $r$, $\delta$, $q$ from \{0.1: 10\}, \{0.8: 8\}, \{0: 20\}, \{60: 100\}, respectively. 
\begin{equation}
    \begin{aligned}
        x=[(y\otimes k_{\sigma})\downarrow_{r}+n_{\delta}]_{\text{JPEG}_{q}}
    \end{aligned}
    \label{formula_degradation}
\end{equation}

    

\begin{figure}[t]
    \centering
    \includegraphics[width=1.0\linewidth]{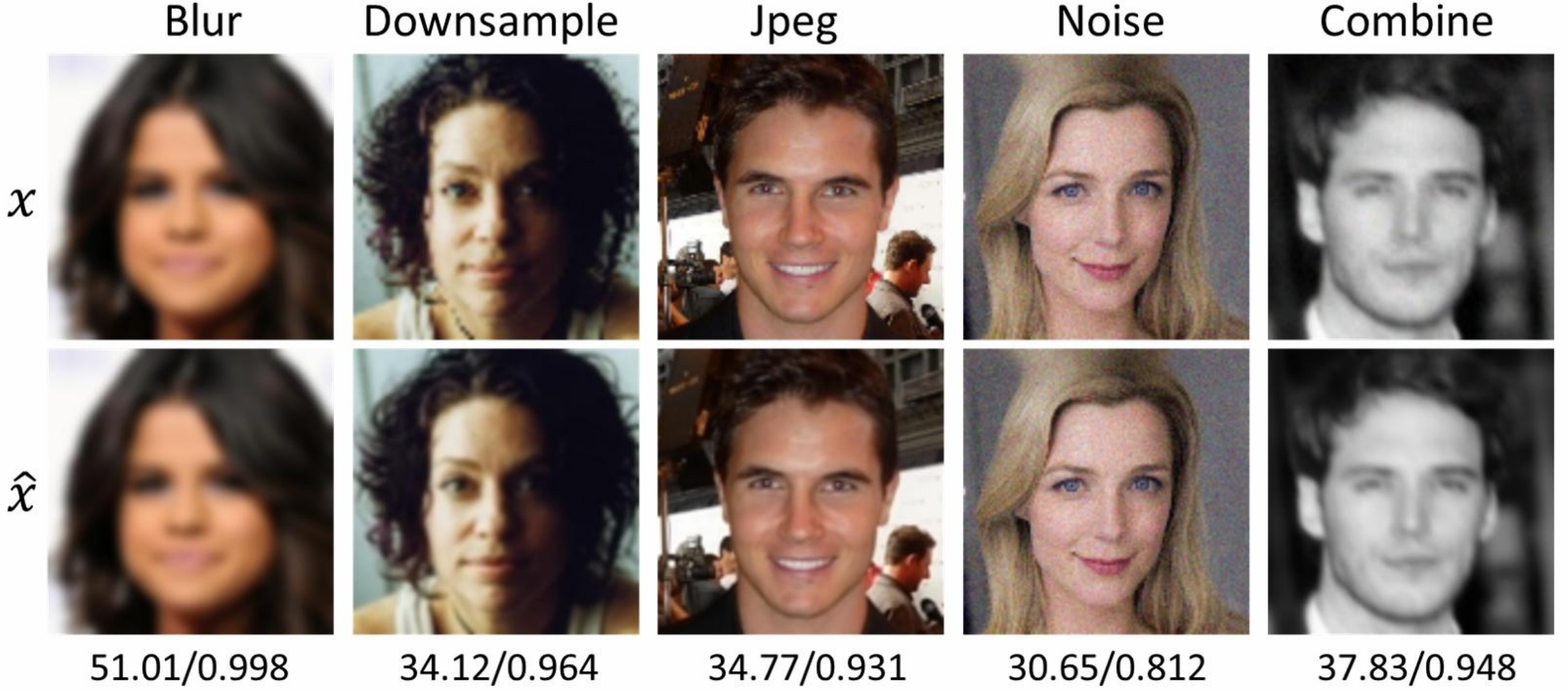}
    \vspace{-6mm}
    \caption{\textbf{Visualization of DR reconstruction}. The output $\hat{x}$ and the original LQ image $x$ present similar degradation, indicating that DR is able to represent sufficient degradation information and reconstruct the degraded image. The bottom row represents the average \textbf{PSNR/SSIM} metrics for 1000 randomly selected samples.} 
    \vspace{-2mm}
    \label{fig:DR_recon}
\end{figure}

\begin{figure}[t]
    \centering
    \includegraphics[width=1.0\linewidth]{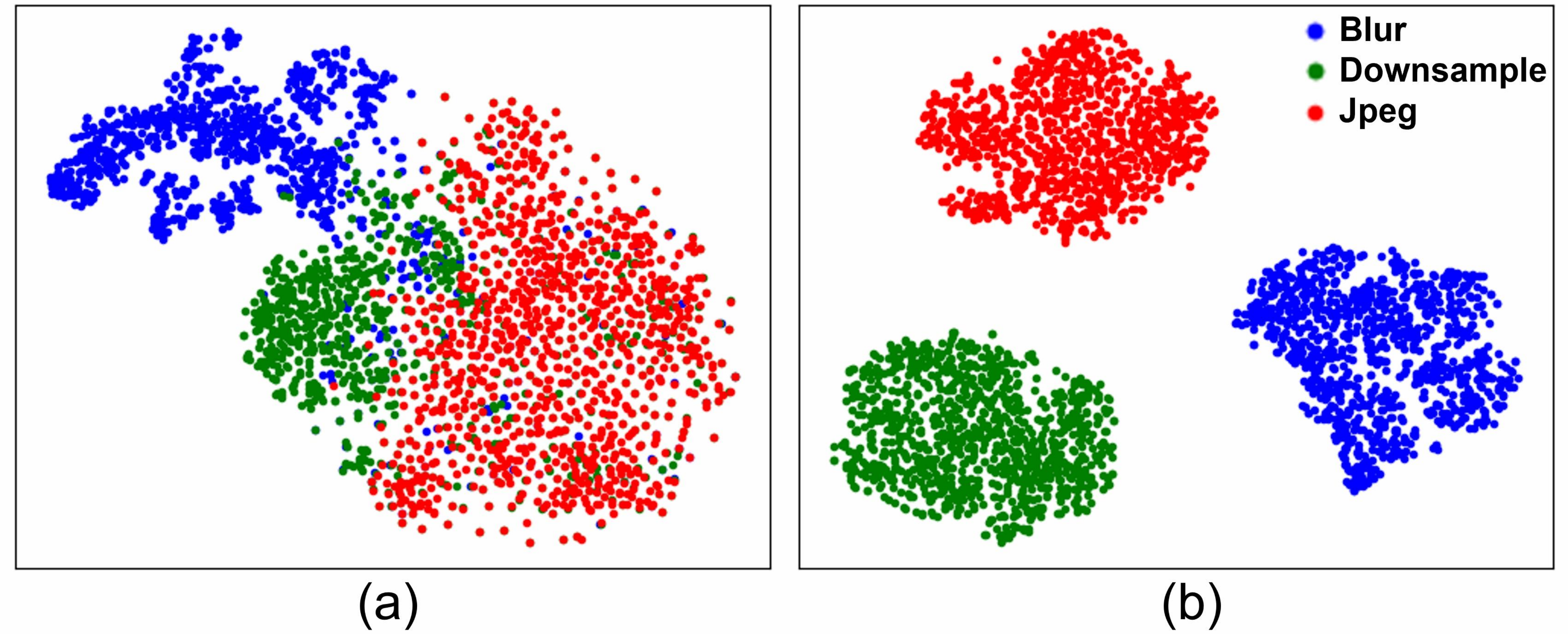}
    \vspace{-6mm}
    \caption{\textbf{Visualization of representations extracted from LQ images} with different types of degradation. \textbf{(a)} illustrates representations under randomly sampled degradation parameters, while \textbf{(b)} depicts representations using extreme degradation parameters across various degradation types.}
    \vspace{-6mm}
    \label{fig:DR_plot}
\end{figure}

\noindent{\textit{\textbf{Testing Datasets.}}}
We select CelebA-Test, LFW~\cite{LFW}, CelebChild~\cite{GFPGAN}, WedPhoto-Test~\cite{GFPGAN} and Wider~\cite{Wider} as the testing datasets. CelebA-Test contains 3,000 HQ images randomly sampled from CelebA-HQ~\cite{Celeba}. The corresponding random LQ images are generated for evaluation using the degradation model specified in Eq.\eqref{formula_degradation}. To assess the generalization capability of our approach, we employed additional real-world datasets including LFW, CelebChild, WedPhoto-Test and Wider. All these datasets have no overlap with our training dataset.

\noindent{\textit{\textbf{Comparison Methods.}}}
We compare DR-BFR with GAN-based, dictionary-based and diffusion-based methods, including GFPGAN~\cite{GFPGAN}, GPEN~\cite{GPEN}, VQFR~\cite{VQFR}, CodeFormer~\cite{Codeformer}, RestoreFormer~\cite{Restoreformer}, DMDNet~\cite{DMDNet}, DifFace~\cite{DifFace}, DiffBIR~\cite{diffbir}, DR2~\cite{dr2} and PGDiff~\cite{pgdiff}.

\noindent{\textit{\textbf{Metrics.}}}
We quantitatively compare differences between our method and state-of-the-art methods using five widely-used metrics, including SSIM~\cite{SSIM}, PSNR, FID~\cite{FID}, NIQE~\cite{NIQE}, and LPIPS~\cite{LPIPS}. Among them, NIQE is a no-reference metric. SSIM and PSNR are pixel-wise similarity measures, while FID, NIQE and LPIPS are perceptual measures.

\subsection{Exploration in Degradation Representation}
We posit that a degradation representation meeting these criteria contains the most complete information, thereby providing sufficient and decoupled conditional guidance for DM. To verify whether the training results of DRM are satisfied, we employ the CelebA-Test dataset and design comparative experiments. Figure~\ref{fig:DR_plot} illustrates the distribution of features extracted by DRM under various degradation types. We randomly selected 1000 images in the CelebA-Test and generated corresponding LQ images through blurring, downsampling and jpeg compression. Subsequently, DRM was used to extract the corresponding features, and $t$-SNE method was applied to visualize the feature distribution. The visualization reveals minimal overlap between the feature distributions of different degradation types, thereby demonstrating the DRM's robust capability to distinguish between various degradations. 

Furthermore, Figure~\ref{fig:DR_recon} illustrates the role of DR in the reconstruction of LQ images. Similarly, 1000 images are randomly selected from CelebA-Test, and random LQ-HQ pairs are obtained according to Eq.\eqref{formula_degradation}. DRM is employed to extract the degradation representation from LQ. Subsequently, degraded images were generated using both HQ images and the extracted DR, simulating original degradation conditions. The similarity between the reconstructed image and the original one is calculated in Figure~\ref{fig:DR_recon}.  The results show that highly similar LQ reconstructed images can be obtained irrespective of degradation types and random combinations of parameters, which demonstrates the maximum coverage of the degradation information by DR.

\begin{figure}[t]
    \centering
    \includegraphics[width=1.0\linewidth]{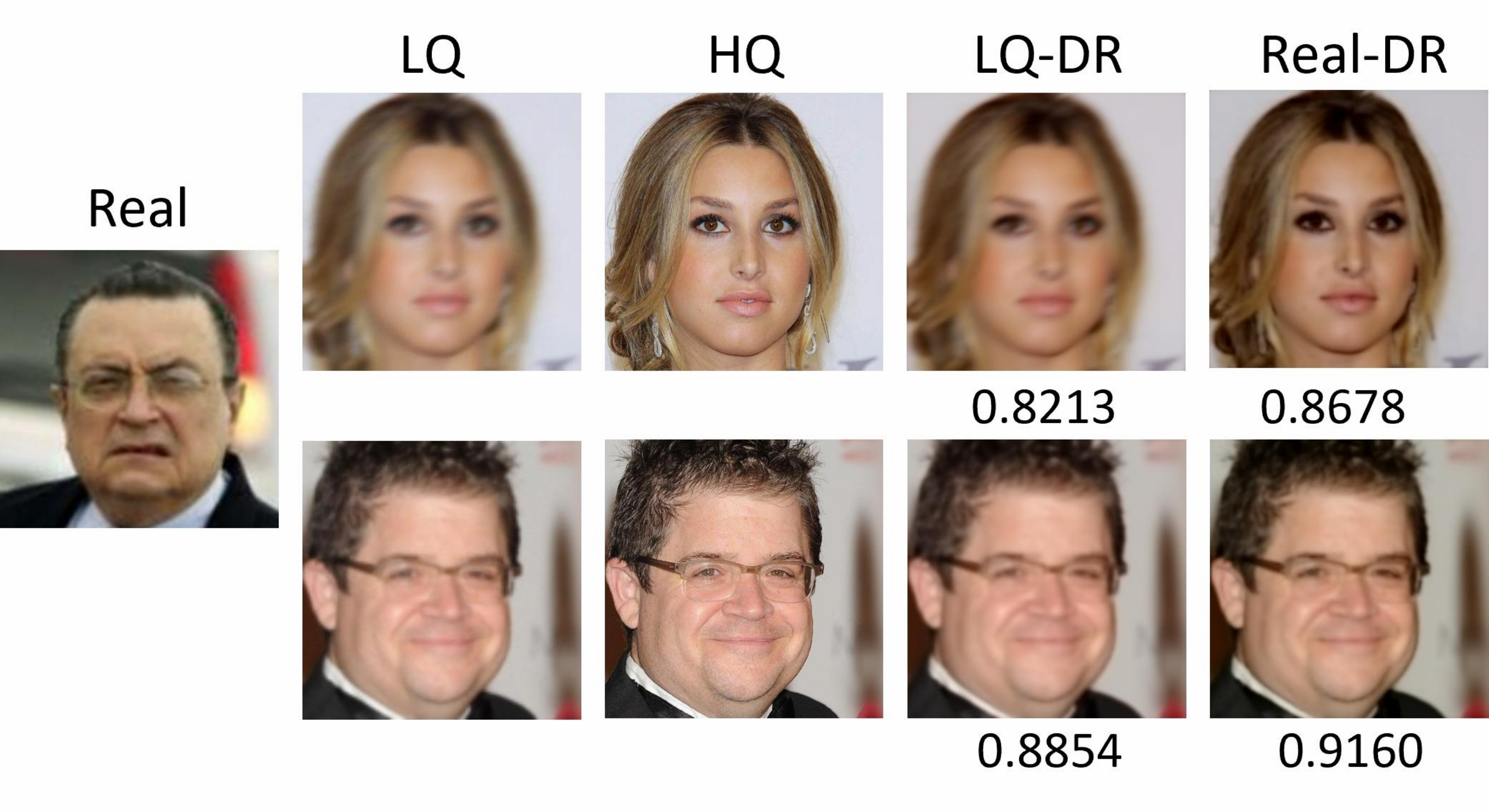}
    \vspace{-8mm}
    \caption{\textbf{Comparison of DR between synthetic and real-world images}. The first column displays real-world image. LQ and HQ images are selected from the generated CelebA-Test dataset to closely match the degradation observed in Real. LQ-DR indicates the DR features used to reconstruct low-quality images from HQ, while Real-DR refers to the DR features of the actual real-world image. Below the picture is the cosine similarity of DR between the current image and the real image.} 
    \vspace{-6mm}
    \label{fig:DR_cos}
\end{figure}

Additionally, to verify the robustness of DR features on real-world datasets, we conduct a comparative analysis as illustrated in Figure~\ref{fig:DR_cos}.  The first column displays real-world images, while LQ and HQ denote two images from the CelebA-Test dataset, selected for their degradation characteristics that closely resemble those of the real images. LQ-DR specifies the DR features extracted from the LQ image, which are used to reconstruct the low-quality image on the HQ image. Real-DR denotes the DR features of the real image. We further compute the cosine similarity between the extracted degradation features and those obtained from the real-world images to validate the reliability of these features. This analytical approach is designed to assess the effectiveness and accuracy of degradation feature extraction and the similarity of the generated images, thereby ensuring the validity of DR features in practical applications.

\subsection{Results on CelebA-Test}
In Figure~\ref{fig:celebA}, we compare visualization results of the proposed DR-BFR and various SOTA methods. Due to space limitation, only the comparative results of the latest methods are presented here, and the complete experimental results are in Appendix. Compared to other methods, DR-BFR not only preserves the original identity and facial expressions of the images, but also generates natural and plausible detail textures. In gray image restoration, DR-BFR excels in maintaining the original color and lighting levels, achieving an optimal balance between restoration accuracy and naturalness. 

Moreover, we quantitatively compare the restoration effects, with the results summarized in Table~\ref{tab:celeba_results}. PSNR and SSIM are pixel-level evaluation metrics, where higher values are often observed in blurrier images, as shown in GAN-based results. Therefore, we consider that achieving normal magnitudes on these metrics is expected. However, FID, NIQE, and LPIPS are feature-level evaluation metrics that are more critical for assessing facial image restoration. The results indicate that DR-BFR outperforms other SOTA methods across almost all these metrics. Notably, FID, which measures the similarity between the data distributions of the restored and original HQ images, achieves the best score with DR-BFR. This suggests that DR-BFR is highly effective in bridging the gap between complex LQ distributions and HQ images distributions.

\begin{figure}[t]
    \centering
    \includegraphics[width=1.0\linewidth]{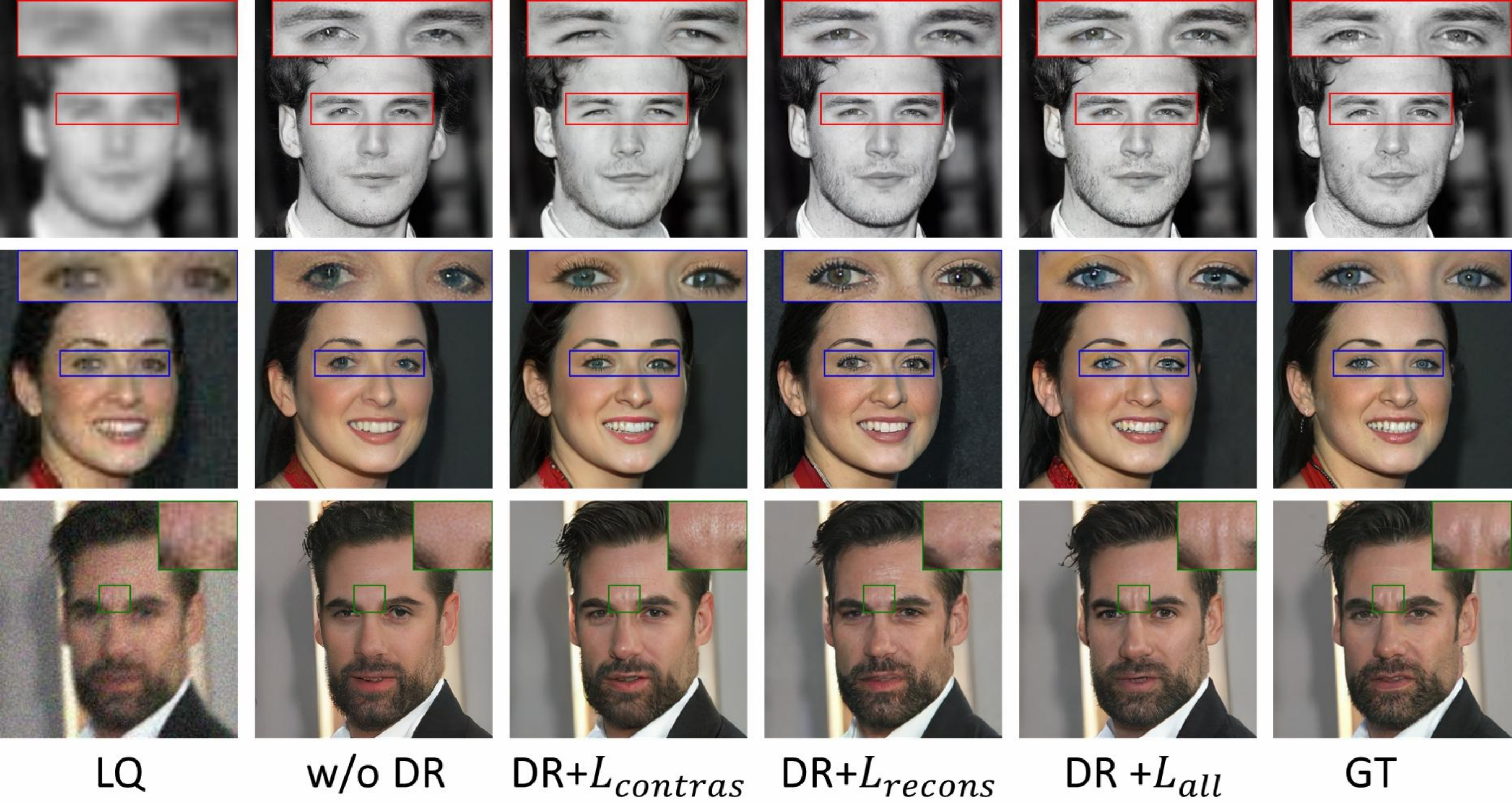}
    \vspace{-6mm}
    \caption{Qualitative comparisons of \textbf{ablation results}. Colored boxes highlight regions where local details are restored under different ablation schemes. \textbf{Zoom in for best view}.}
    \vspace{-3mm}
    \label{fig:ablation}
\end{figure}

\begin{table}[t]
\setlength{\abovecaptionskip}{0.cm}
\setlength{\belowcaptionskip}{-0.cm}
    \begin{center}
    \setlength\tabcolsep{4.0pt}
    \begin{tabular}{l|cc|ccc}
    \bottomrule
    Method & SSIM$\uparrow$ & PSNR$\uparrow$ & FID$\downarrow$ & NIQE$\downarrow$ & LPIPS$\downarrow$ \\
    \hline
     DR-None           & 0.6304&	24.522&	20.103&	5.8928&	0.2900 \\
     DR-CL & \textbf{0.6635}&	\textbf{25.078}& 18.730& 5.6396&	0.2642\\
     DR-REC  & 0.6521&	24.698&	19.644&	5.3073&	0.2653 \\
     \textbf{DR-ALL}     & 0.6429&	24.549&	\textbf{13.686}&	\textbf{5.0113}&	\textbf{0.2499} \\
    \toprule
    \end{tabular}
    \end{center}
    \vspace{-2mm}
    \caption{\textbf{Ablation study} on CelebA-Test. \emph{DR-None}: without DR inject in LDM; \emph{DR-CL}: DRM trained solely with contrastive learning; \emph{DR-REC}:DRM trained solely with LQ reconstruction; \textbf{\emph{DR-ALL}}: full modules in DR-BFR.}
    \vspace{-6mm}
    \label{ablationResult}
\end{table}

\subsection{Results on Real World Dataset}
\vspace{-2mm}
LQ images in real-world datasets are directly acquired from real-world scenes, where the causes of degradation are highly complex and varied. Consequently, corresponding HQ image pairs are not available. Therefore, the results on real-world datasets can only be quantitatively evaluated using non-reference metrics such as NIQE and FID, and the generalization ability of various methods in complex environments can be judged by observing the visualized results. 

Figure~\ref{fig:real} illustrates the visualization results of DR-BFR compared with several SOTA methods. The inference images obtained by GAN-based methods lose significant detail texture and often confuse facial texture information with degradation noise. Conversely, dictionary-based methods generate overly smooth images that lack the naturalness observed in real-world scenes. DM-based SOTA mehods are capable of generating more detailed textures; however, their performance deteriorates on severely degraded images due to their limited ability to discriminate between degraded features. In contrast, our DR-BFR achieves more reasonable and detailed results by leveraging the perceived DR from severely degraded inputs, without being affected by excessive ambiguity. Table~\ref{tab:realworld} presents the comparison results of non-reference metrics, and DR-BFR significantly outperforms the other methods in both NIQE and FID.


\subsection{Ablation Studies}\label{sec:ablation}
\vspace{-2mm}
To verify the effectiveness of each module design within DR-BFR, we present the quantitative results from a series of ablation experiments in Table~\ref{ablationResult}. Initially, we configured the unconditional LDM to directly embed the LQ image as guidance. We also compare schemes that utilize only contrastive learning or only LQ reconstruction loss for training feature extractors in degenerate representations, denoted as DR-CL and DR-REC, respectively. These three configurations are compared with the complete DR-BFR to assess the effectiveness of DRM in guiding LDM. 
Figure~\ref{fig:ablation} qualitatively compares the results of the four schemes in the CelebA-Test. LDM without DR guidance exhibits poor restoration performance on LQ images with varying levels of degradation; detailed textures in the inferred images are very unclear and often appear merged. This issue is particularly pronounced in LQ images with severe degradation, resulting in highly unnatural outcomes. Both DR-CL and DR-REC perform slightly worse than DR-BFR in quantitative metrics, and exhibit artifacts that cannot be identified and corrected. This underscores the necessity of combining both training approaches to achieve optimal performance.
\vspace{-1mm}
\section{Discussion\ }
\vspace{-2mm}
\noindent{\textit{\textbf{Advantages.}}}
We design a content-independent DR that maximizes the capture of degradation, and jointly train the feature extractor by combining unsupervised contrastive learning with LQ reconstruction. Furthermore, we demonstrate that degenerate representation plays a crucial role in the diffusion model-based restoration framework, particularly in addressing blind inverse problems.

\noindent{\textit{\textbf{Limitations.}}}
The figure in Appendix illustrates failure cases of DR-BFR. When real-world images contain watermarks, the inference results of DR-BFR still exhibit residual artifacts. This issue primarily arises because the model is trained on synthetically generated degradation datasets, which do not encompass the complexity of real-world degradations. Our future research aims to leverage unsupervised learning in DRM to enhance their applicability on real-world LQ samples within DM-based method.

\section{Conclusion\ }
\vspace{-2mm}
This paper presents the methodological framework of DR-BFR, which enhances the restoration performance of DM by leveraging DR. The core of our approach involves decoupling degraded and diverse LQ face images in advance to extract the DR features as the degradation prompt guidance that are independent of image content while maximizing degradation information, and then guide the inverse diffusion generation process of LDM with LQ inputs. DR-BFR demonstrates superior results on both synthetically degraded and real-world datasets, outperforming previous methods in terms of the naturalness and fidelity of the restored images.
\clearpage
{

    \small
    \bibliographystyle{ieeenat_fullname}
    \bibliography{main}

\begin{thebibliography}{55}
\providecommand{\natexlab}[1]{#1}
\providecommand{\url}[1]{\texttt{#1}}
\expandafter\ifx\csname urlstyle\endcsname\relax
  \providecommand{\doi}[1]{doi: #1}\else
  \providecommand{\doi}{doi: \begingroup \urlstyle{rm}\Url}\fi

\bibitem[Chen et~al.(2021)Chen, Li, Yang, Lin, Zhang, and Wong]{PSFRGAN}
Chaofeng Chen, Xiaoming Li, Lingbo Yang, Xianhui Lin, Lei Zhang, and Kwan-Yee~K Wong.
\newblock Progressive semantic-aware style transformation for blind face restoration.
\newblock In \emph{CVPR}, pages 11896--11905, 2021.

\bibitem[Chen et~al.(2023)Chen, Zhang, Gu, Yuan, Kong, Chen, and Yang]{prompt_sr}
Zheng Chen, Yulun Zhang, Jinjin Gu, Xin Yuan, Linghe Kong, Guihai Chen, and Xiaokang Yang.
\newblock Image super-resolution with text prompt diffusion.
\newblock \emph{arXiv preprint arXiv:2311.14282}, 2023.

\bibitem[Chng et~al.(2024)Chng, Saratchandran, and Lucey]{Implicit_da}
Shin-Fang Chng, Hemanth Saratchandran, and Simon Lucey.
\newblock Preconditioners for the stochastic training of implicit neural representations.
\newblock \emph{arXiv preprint arXiv:2402.08784}, 2024.

\bibitem[Delbracio and Milanfar(2023)]{InDI}
Mauricio Delbracio and Peyman Milanfar.
\newblock Inversion by direct iteration: An alternative to denoising diffusion for image restoration.
\newblock \emph{arXiv preprint arXiv:2303.11435}, 2023.

\bibitem[Esser et~al.(2021)Esser, Rombach, and Ommer]{vqgan}
Patrick Esser, Robin Rombach, and Bjorn Ommer.
\newblock Taming transformers for high-resolution image synthesis.
\newblock In \emph{Proceedings of the IEEE/CVF conference on computer vision and pattern recognition}, pages 12873--12883, 2021.

\bibitem[Gu et~al.(2022)Gu, Wang, Xie, Dong, Li, Shan, and Cheng]{VQFR}
Yuchao Gu, Xintao Wang, Liangbin Xie, Chao Dong, Gen Li, Ying Shan, and Ming-Ming Cheng.
\newblock Vqfr: Blind face restoration with vector-quantized dictionary and parallel decoder.
\newblock In \emph{ECCV}, pages 126--143. Springer, 2022.

\bibitem[Heusel et~al.(2017)Heusel, Ramsauer, Unterthiner, Nessler, and Hochreiter]{FID}
Martin Heusel, Hubert Ramsauer, Thomas Unterthiner, Bernhard Nessler, and Sepp Hochreiter.
\newblock Gans trained by a two time-scale update rule converge to a local nash equilibrium.
\newblock In \emph{NeurIPS}, 2017.

\bibitem[Ho et~al.(2020)Ho, Jain, and Abbeel]{DDPM}
Jonathan Ho, Ajay Jain, and Pieter Abbeel.
\newblock Denoising diffusion probabilistic models.
\newblock In \emph{NeurIPS}, pages 6840--6851, 2020.

\bibitem[Huang et~al.(2008)Huang, Mattar, Berg, and Learned-Miller]{LFW}
Gary~B Huang, Marwan Mattar, Tamara Berg, and Eric Learned-Miller.
\newblock Labeled faces in the wild: A database for studying face recognition in unconstrained environments.
\newblock In \emph{Workshop on faces in'Real-Life'Images: detection, alignment, and recognition}, 2008.

\bibitem[Jiang et~al.(2023)Jiang, Zhang, Xue, and Gu]{allinone}
Yitong Jiang, Zhaoyang Zhang, Tianfan Xue, and Jinwei Gu.
\newblock Autodir: Automatic all-in-one image restoration with latent diffusion.
\newblock \emph{arXiv preprint arXiv:2310.10123}, 2023.

\bibitem[Karras et~al.(2018)Karras, Aila, Laine, and Lehtinen]{Celeba}
Tero Karras, Timo Aila, Samuli Laine, and Jaakko Lehtinen.
\newblock Progressive growing of gans for improved quality, stability, and variation.
\newblock In \emph{ICLR}, 2018.

\bibitem[Karras et~al.(2019)Karras, Laine, and Aila]{styleGAN}
Tero Karras, Samuli Laine, and Timo Aila.
\newblock A style-based generator architecture for generative adversarial networks.
\newblock In \emph{CVPR}, pages 4401--4410, 2019.

\bibitem[Kawar et~al.(2022)Kawar, Elad, Ermon, and Song]{DDRM}
Bahjat Kawar, Michael Elad, Stefano Ermon, and Jiaming Song.
\newblock Denoising diffusion restoration models.
\newblock \emph{Advances in Neural Information Processing Systems}, 35:\penalty0 23593--23606, 2022.

\bibitem[Li et~al.(2022{\natexlab{a}})Li, Liu, Hu, Wu, Lv, and Peng]{AirNet}
Boyun Li, Xiao Liu, Peng Hu, Zhongqin Wu, Jiancheng Lv, and Xi Peng.
\newblock All-in-one image restoration for unknown corruption.
\newblock In \emph{Proceedings of the IEEE/CVF conference on computer vision and pattern recognition}, pages 17452--17462, 2022{\natexlab{a}}.

\bibitem[Li et~al.(2020)Li, Chen, Zhou, Lin, Zuo, and Zhang]{DFDNet}
Xiaoming Li, Chaofeng Chen, Shangchen Zhou, Xianhui Lin, Wangmeng Zuo, and Lei Zhang.
\newblock Blind face restoration via deep multi-scale component dictionaries.
\newblock pages 399--415, 2020.

\bibitem[Li et~al.(2022{\natexlab{b}})Li, Zhang, Zhou, Zhang, and Zuo]{DMDNet}
Xiaoming Li, Shiguang Zhang, Shangchen Zhou, Lei Zhang, and Wangmeng Zuo.
\newblock Learning dual memory dictionaries for blind face restoration.
\newblock \emph{IEEE Transactions on Pattern Analysis and Machine Intelligence}, 45\penalty0 (5):\penalty0 5904--5917, 2022{\natexlab{b}}.

\bibitem[Li et~al.(2024)Li, Li, Jin, Lan, Zhu, Ren, and Chen]{prompt_ucip}
Xin Li, Bingchen Li, Yeying Jin, Cuiling Lan, Hanxin Zhu, Yulin Ren, and Zhibo Chen.
\newblock Ucip: A universal framework for compressed image super-resolution using dynamic prompt.
\newblock \emph{arXiv preprint arXiv:2407.13108}, 2024.

\bibitem[Lin et~al.(2023)Lin, He, Chen, Lyu, Fei, Dai, Ouyang, Qiao, and Dong]{diffbir}
Xinqi Lin, Jingwen He, Ziyan Chen, Zhaoyang Lyu, Ben Fei, Bo Dai, Wanli Ouyang, Yu Qiao, and Chao Dong.
\newblock Diffbir: Towards blind image restoration with generative diffusion prior.
\newblock \emph{arXiv preprint arXiv:2308.15070}, 2023.

\bibitem[Liu et~al.(2024)Liu, Wang, Zhang, Xue, Zhou, and Guo]{ReDSR}
Hongda Liu, Longguang Wang, Ye Zhang, Kaiwen Xue, Shunbo Zhou, and Yulan Guo.
\newblock Preserving full degradation details for blind image super-resolution.
\newblock \emph{arXiv preprint arXiv:2407.01299}, 2024.

\bibitem[Mahara et~al.(2024)Mahara, Rishe, and Deng]{GAN_I2I}
Arpan Mahara, Naphtali~D Rishe, and Liangdong Deng.
\newblock The dawn of kan in image-to-image (i2i) translation: Integrating kolmogorov-arnold networks with gans for unpaired i2i translation.
\newblock \emph{arXiv preprint arXiv:2408.08216}, 2024.

\bibitem[Menon et~al.(2020)Menon, Damian, Hu, Ravi, and Rudin]{PULSE}
Sachit Menon, Alexandru Damian, Shijia Hu, Nikhil Ravi, and Cynthia Rudin.
\newblock Pulse: Self-supervised photo upsampling via latent space exploration of generative models.
\newblock In \emph{CVPR}, pages 2437--2445, 2020.

\bibitem[Mirza(2014)]{gannet}
Mehdi Mirza.
\newblock Conditional generative adversarial nets.
\newblock \emph{arXiv preprint arXiv:1411.1784}, 2014.

\bibitem[Mittal et~al.(2012)Mittal, Soundararajan, and Bovik]{NIQE}
Anish Mittal, Rajiv Soundararajan, and Alan~C Bovik.
\newblock Making a “completely blind” image quality analyzer.
\newblock \emph{IEEE Signal processing letters}, 20\penalty0 (3):\penalty0 209--212, 2012.

\bibitem[Nichol and Dhariwal(2021)]{IprovedDDPM}
Alexander~Quinn Nichol and Prafulla Dhariwal.
\newblock Improved denoising diffusion probabilistic models.
\newblock In \emph{ICML}, pages 8162--8171, 2021.

\bibitem[Preechakul et~al.(2022)Preechakul, Chatthee, Wizadwongsa, and Suwajanakorn]{decodable_da}
Konpat Preechakul, Nattanat Chatthee, Suttisak Wizadwongsa, and Supasorn Suwajanakorn.
\newblock Diffusion autoencoders: Toward a meaningful and decodable representation.
\newblock In \emph{Proceedings of the IEEE/CVF conference on computer vision and pattern recognition}, pages 10619--10629, 2022.

\bibitem[Rombach et~al.(2022)Rombach, Blattmann, Lorenz, Esser, and Ommer]{ldm}
Robin Rombach, Andreas Blattmann, Dominik Lorenz, Patrick Esser, and Bj{\"o}rn Ommer.
\newblock High-resolution image synthesis with latent diffusion models.
\newblock In \emph{Proceedings of the IEEE/CVF conference on computer vision and pattern recognition}, pages 10684--10695, 2022.

\bibitem[Saharia et~al.(2022{\natexlab{a}})Saharia, Chan, Saxena, Li, Whang, Denton, Ghasemipour, Gontijo~Lopes, Karagol~Ayan, Salimans, et~al.]{imagen}
Chitwan Saharia, William Chan, Saurabh Saxena, Lala Li, Jay Whang, Emily~L Denton, Kamyar Ghasemipour, Raphael Gontijo~Lopes, Burcu Karagol~Ayan, Tim Salimans, et~al.
\newblock Photorealistic text-to-image diffusion models with deep language understanding.
\newblock \emph{Advances in neural information processing systems}, 35:\penalty0 36479--36494, 2022{\natexlab{a}}.

\bibitem[Saharia et~al.(2022{\natexlab{b}})Saharia, Ho, Chan, Salimans, Fleet, and Norouzi]{SR3}
Chitwan Saharia, Jonathan Ho, William Chan, Tim Salimans, David~J Fleet, and Mohammad Norouzi.
\newblock Image super-resolution via iterative refinement.
\newblock \emph{IEEE Transactions on Pattern Analysis and Machine Intelligence}, 45\penalty0 (4):\penalty0 4713--4726, 2022{\natexlab{b}}.

\bibitem[Simonyan(2014)]{vgg}
Karen Simonyan.
\newblock Very deep convolutional networks for large-scale image recognition.
\newblock \emph{arXiv preprint arXiv:1409.1556}, 2014.

\bibitem[Song et~al.(2020)Song, Meng, and Ermon]{DDIM}
Jiaming Song, Chenlin Meng, and Stefano Ermon.
\newblock Denoising diffusion implicit models.
\newblock \emph{arXiv preprint arXiv:2010.02502}, 2020.

\bibitem[Tie et~al.(2024)Tie, Wei, Wang, Wu, Yuan, Zhang, Jia, Zhao, Gan, and Ding]{o2v_da}
Muer Tie, Julong Wei, Zhengjun Wang, Ke Wu, Shansuai Yuan, Kaizhao Zhang, Jie Jia, Jieru Zhao, Zhongxue Gan, and Wenchao Ding.
\newblock O2v-mapping: Online open-vocabulary mapping with neural implicit representation.
\newblock \emph{arXiv preprint arXiv:2404.06836}, 2024.

\bibitem[Wang et~al.(2023{\natexlab{a}})Wang, Pan, Wang, Dong, Wang, Ju, and Chen]{promptrestorer}
Cong Wang, Jinshan Pan, Wei Wang, Jiangxin Dong, Mengzhu Wang, Yakun Ju, and Junyang Chen.
\newblock Promptrestorer: A prompting image restoration method with degradation perception.
\newblock \emph{Advances in Neural Information Processing Systems}, 36:\penalty0 8898--8912, 2023{\natexlab{a}}.

\bibitem[Wang et~al.(2021{\natexlab{a}})Wang, Wang, Dong, Xu, Yang, An, and Guo]{dasr2}
Longguang Wang, Yingqian Wang, Xiaoyu Dong, Qingyu Xu, Jungang Yang, Wei An, and Yulan Guo.
\newblock Unsupervised degradation representation learning for blind super-resolution.
\newblock In \emph{CVPR}, 2021{\natexlab{a}}.

\bibitem[Wang et~al.(2024)Wang, Lu, Zhang, Luo, Kim, Lu, Li, and Yang]{promptrr}
Tao Wang, Wanglong Lu, Kaihao Zhang, Wenhan Luo, Tae-Kyun Kim, Tong Lu, Hongdong Li, and Ming-Hsuan Yang.
\newblock Promptrr: Diffusion models as prompt generators for single image reflection removal.
\newblock \emph{arXiv preprint arXiv:2402.02374}, 2024.

\bibitem[Wang et~al.(2021{\natexlab{b}})Wang, Li, Zhang, and Shan]{GFPGAN}
Xintao Wang, Yu Li, Honglun Zhang, and Ying Shan.
\newblock Towards real-world blind face restoration with generative facial prior.
\newblock pages 9168--9178, 2021{\natexlab{b}}.

\bibitem[Wang et~al.(2023{\natexlab{b}})Wang, Yu, and Zhang]{DDNM}
Yinhuai Wang, Jiwen Yu, and Jian Zhang.
\newblock Zero-shot image restoration using denoising diffusion null-space model.
\newblock In \emph{The Eleventh International Conference on Learning Representations}, 2023{\natexlab{b}}.

\bibitem[Wang et~al.(2004)Wang, Bovik, Sheikh, and Simoncelli]{SSIM}
Zhou Wang, Alan~C Bovik, Hamid~R Sheikh, and Eero~P Simoncelli.
\newblock Image quality assessment: from error visibility to structural similarity.
\newblock \emph{IEEE transactions on image processing}, 13\penalty0 (4):\penalty0 600--612, 2004.

\bibitem[Wang et~al.(2022)Wang, Zhang, Chen, Wang, and Luo]{Restoreformer}
Zhouxia Wang, Jiawei Zhang, Runjian Chen, Wenping Wang, and Ping Luo.
\newblock Restoreformer: High-quality blind face restoration from undegraded key-value pairs.
\newblock In \emph{Proceedings of the IEEE/CVF conference on computer vision and pattern recognition}, pages 17512--17521, 2022.

\bibitem[Wang et~al.(2023{\natexlab{c}})Wang, Zhang, Zhang, Zheng, Zhou, Zhang, and Wang]{dr2}
Zhixin Wang, Ziying Zhang, Xiaoyun Zhang, Huangjie Zheng, Mingyuan Zhou, Ya Zhang, and Yanfeng Wang.
\newblock Dr2: Diffusion-based robust degradation remover for blind face restoration.
\newblock In \emph{Proceedings of the IEEE/CVF Conference on Computer Vision and Pattern Recognition}, pages 1704--1713, 2023{\natexlab{c}}.

\bibitem[Wei et~al.(2021)Wei, Gu, Li, Timofte, Jin, and Song]{dasr}
Yunxuan Wei, Shuhang Gu, Yawei Li, Radu Timofte, Longcun Jin, and Hengjie Song.
\newblock Unsupervised real-world image super resolution via domain-distance aware training.
\newblock In \emph{Proceedings of the IEEE/CVF conference on computer vision and pattern recognition}, pages 13385--13394, 2021.

\bibitem[Wolf et~al.(2021)Wolf, Lugmayr, Danelljan, Van~Gool, and Timofte]{deflow_da}
Valentin Wolf, Andreas Lugmayr, Martin Danelljan, Luc Van~Gool, and Radu Timofte.
\newblock Deflow: Learning complex image degradations from unpaired data with conditional flows.
\newblock In \emph{Proceedings of the IEEE/CVF Conference on Computer Vision and Pattern Recognition}, pages 94--103, 2021.

\bibitem[Xia et~al.(2023)Xia, Zhang, Wang, Wang, Wu, Tian, Yang, and Van~Gool]{diffir}
Bin Xia, Yulun Zhang, Shiyin Wang, Yitong Wang, Xinglong Wu, Yapeng Tian, Wenming Yang, and Luc Van~Gool.
\newblock Diffir: Efficient diffusion model for image restoration.
\newblock In \emph{Proceedings of the IEEE/CVF International Conference on Computer Vision}, pages 13095--13105, 2023.

\bibitem[Yang et~al.(2024)Yang, Zhou, Tao, and Loy]{pgdiff}
Peiqing Yang, Shangchen Zhou, Qingyi Tao, and Chen~Change Loy.
\newblock Pgdiff: Guiding diffusion models for versatile face restoration via partial guidance.
\newblock \emph{Advances in Neural Information Processing Systems}, 36, 2024.

\bibitem[Yang et~al.(2016)Yang, Luo, Loy, and Tang]{Wider}
Shuo Yang, Ping Luo, Chen-Change Loy, and Xiaoou Tang.
\newblock Wider face: A face detection benchmark.
\newblock In \emph{Proceedings of the IEEE conference on computer vision and pattern recognition}, pages 5525--5533, 2016.

\bibitem[Yang et~al.(2021)Yang, Ren, Xie, and Zhang]{GPEN}
Tao Yang, Peiran Ren, Xuansong Xie, and Lei Zhang.
\newblock Gan prior embedded network for blind face restoration in the wild.
\newblock In \emph{CVPR}, pages 672--681, 2021.

\bibitem[Yao et~al.(2023)Yao, Xu, Guan, Huang, and Xiong]{NDR}
Mingde Yao, Ruikang Xu, Yuanshen Guan, Jie Huang, and Zhiwei Xiong.
\newblock Neural degradation representation learning for all-in-one image restoration.
\newblock \emph{arXiv preprint arXiv:2310.12848}, 2023.

\bibitem[Yu et~al.(2024)Yu, Peng, Zhu, Zhang, Tian, and Lei]{prompt_seek}
Chang Yu, Junran Peng, Xiangyu Zhu, Zhaoxiang Zhang, Qi Tian, and Zhen Lei.
\newblock Seek for incantations: Towards accurate text-to-image diffusion synthesis through prompt engineering.
\newblock \emph{arXiv preprint arXiv:2401.06345}, 2024.

\bibitem[Yue and Loy(2022)]{DifFace}
Zongsheng Yue and Chen~Change Loy.
\newblock Difface: Blind face restoration with diffused error contraction.
\newblock 2022.

\bibitem[Yue et~al.(2024)Yue, Wang, and Loy]{res_shift}
Zongsheng Yue, Jianyi Wang, and Chen~Change Loy.
\newblock Efficient diffusion model for image restoration by residual shifting.
\newblock \emph{arXiv preprint arXiv:2403.07319}, 2024.

\bibitem[Zamfir et~al.(2024)Zamfir, Wu, Mehta, Paudel, Zhang, and Timofte]{DaAIR}
Eduard Zamfir, Zongwei Wu, Nancy Mehta, Danda~Dani Paudel, Yulun Zhang, and Radu Timofte.
\newblock Efficient degradation-aware any image restoration.
\newblock \emph{arXiv preprint arXiv:2405.15475}, 2024.

\bibitem[Zhang et~al.(2022)Zhang, Gu, Zhang, Bao, Chen, Wen, Wang, and Guo]{styleswinGAN}
Bowen Zhang, Shuyang Gu, Bo Zhang, Jianmin Bao, Dong Chen, Fang Wen, Yong Wang, and Baining Guo.
\newblock Styleswin: Transformer-based gan for high-resolution image generation.
\newblock In \emph{Proceedings of the IEEE/CVF conference on computer vision and pattern recognition}, pages 11304--11314, 2022.

\bibitem[Zhang et~al.(2018{\natexlab{a}})Zhang, Zuo, and Zhang]{degradation2}
Kai Zhang, Wangmeng Zuo, and Lei Zhang.
\newblock Learning a single convolutional super-resolution network for multiple degradations.
\newblock In \emph{CVPR}, pages 3262--3271, 2018{\natexlab{a}}.

\bibitem[Zhang et~al.(2018{\natexlab{b}})Zhang, Isola, Efros, Shechtman, and Wang]{LPIPS}
Richard Zhang, Phillip Isola, Alexei~A Efros, Eli Shechtman, and Oliver Wang.
\newblock The unreasonable effectiveness of deep features as a perceptual metric.
\newblock In \emph{CVPR}, pages 586--595, 2018{\natexlab{b}}.

\bibitem[Zheng et~al.(2024)Zheng, Wu, Yang, Zhang, Hu, and Zheng]{selective}
Dian Zheng, Xiao-Ming Wu, Shuzhou Yang, Jian Zhang, Jian-Fang Hu, and Wei-Shi Zheng.
\newblock Selective hourglass mapping for universal image restoration based on diffusion model.
\newblock In \emph{Proceedings of the IEEE/CVF Conference on Computer Vision and Pattern Recognition}, pages 25445--25455, 2024.

\bibitem[Zhou et~al.(2022)Zhou, Chan, Li, and Loy]{Codeformer}
Shangchen Zhou, Kelvin Chan, Chongyi Li, and Chen~Change Loy.
\newblock Towards robust blind face restoration with codebook lookup transformer.
\newblock \emph{Advances in Neural Information Processing Systems}, 35:\penalty0 30599--30611, 2022.

\end{thebibliography}
}


\end{document}